\documentclass{article}

\usepackage{microtype}
\usepackage{graphicx}
\usepackage{booktabs} 
\usepackage{subcaption}

\usepackage{hyperref}

\usepackage[accepted]{icml2021}

\icmltitlerunning{Reimagining GNN Explanations}

\begin{document}

\twocolumn[
\icmltitle{Reimagining GNN Explanations with ideas from Tabular Data}




\begin{icmlauthorlist}
\icmlauthor{Anjali Singh}{mahe}
\icmlauthor{Shamanth R Nayak K}{mahe}
\icmlauthor{Balaji Ganesan}{irl}
\end{icmlauthorlist}

\icmlaffiliation{mahe}{Manipal Institute of Technology, Manipal, India}
\icmlaffiliation{irl}{IBM Research, Bengaluru, India}

\icmlcorrespondingauthor{Balaji Ganesan}{bganesa1@in.ibm.com}

\icmlkeywords{Machine Learning, ICML}

\vskip 0.3in
]



\printAffiliationsAndNotice{}  

\begin{abstract}
Explainability techniques for Graph Neural Networks still have a long way to go compared to explanations available for both neural and decision decision tree-based models trained on tabular data. Using a task that straddles both graphs and tabular data, namely Entity Matching, we comment on key aspects of explainability that are missing in GNN model explanations.
\end{abstract}

\section{Introduction}

Explainability techniques for Graph Neural Networks are an active area of research. Just as new methods are being proposed, it could be useful to take a step back and compare these methods with the progress made in other areas of machine learning.

Over the years, deep neural networks models have benefited from applying ideas from one domain to another. Explainability techniques too have been adopted across domains, though explanations in some domains (like tabular data) might be inherently easier than others (like graphs).

Entity matching is the task of predicting if two entities in a database belong to the same real-world entity. This task has a number of enterprise applications in sales recommendations, financial services and customer support. People and organization entities are of particular interest though entity matching also applies to products, locations, events, and even abstract ideas like compliance clauses and legal cases.

\begin{figure}[htb]
    \centering
    \includegraphics[width=0.9\columnwidth]{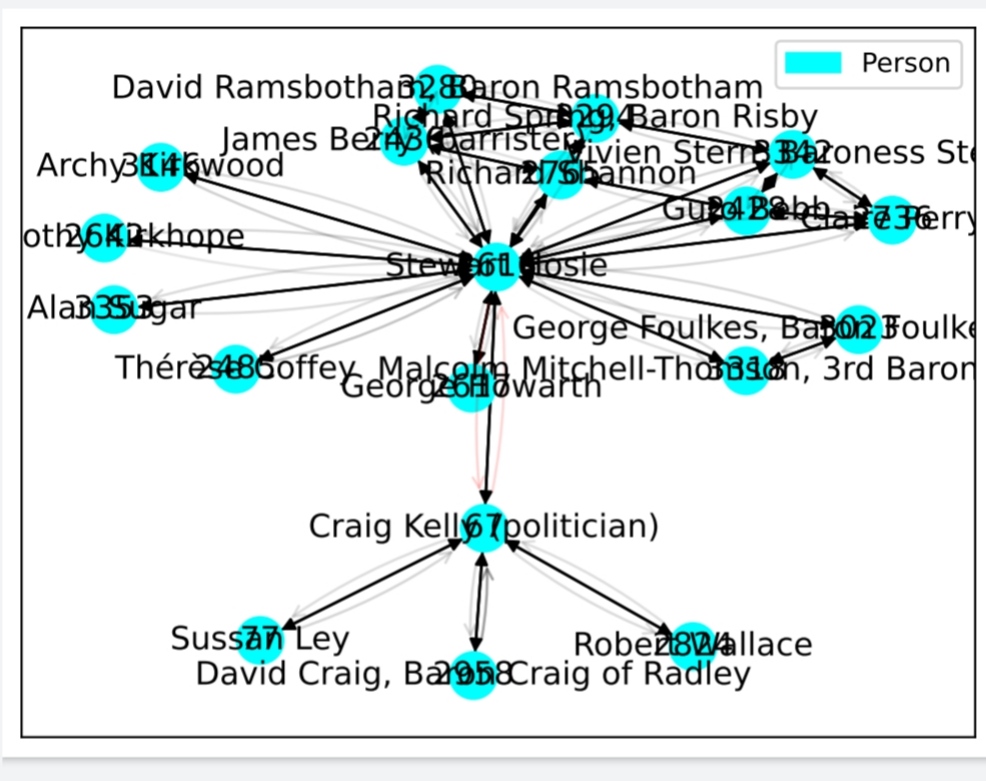}
    \caption{A typical subgraph explanation from GNN explainability solutions, which are hard to understand for end users}
    \label{fig:gnn_explainer_example}
\end{figure}

A somewhat unique aspect of entity matching is that it could be treated both as a task in tabular data and graphs. Hence using entity matching as our task, we explored explainability techniques in both tabular data and graph neural networks, and have identified some key aspects of explainability that are missing in GNN model explanations. In particular, we looked at very popular explainability techniques in tabular data, namely LIME \cite{ribeiro2016should}, SHAP \cite{lundberg2017unified}, Anchors \cite{ribeiro2018anchors} and TabNET \cite{TabNet}, that have been widely adopted in the industry and not just recent advances in research.

\begin{figure*}[!htb]
\begin{subfigure}{0.48\textwidth}
    \centering
    \includegraphics[height=6cm, width=\columnwidth]{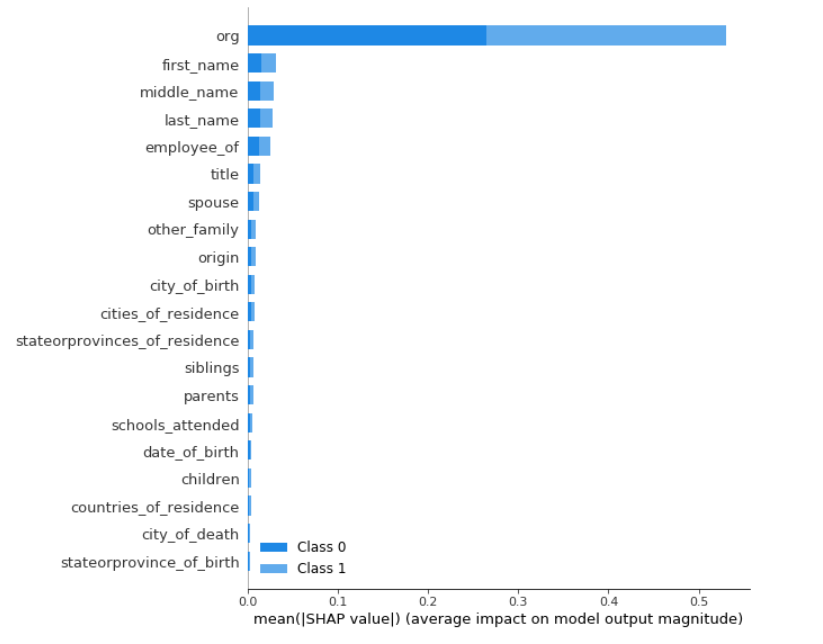}
    \caption{Feature importance using SHAP}
    \label{fig:shap_example}
\end{subfigure}
\begin{subfigure}{0.48\textwidth}
    \centering
    \includegraphics[height=6cm, width=0.5\columnwidth]{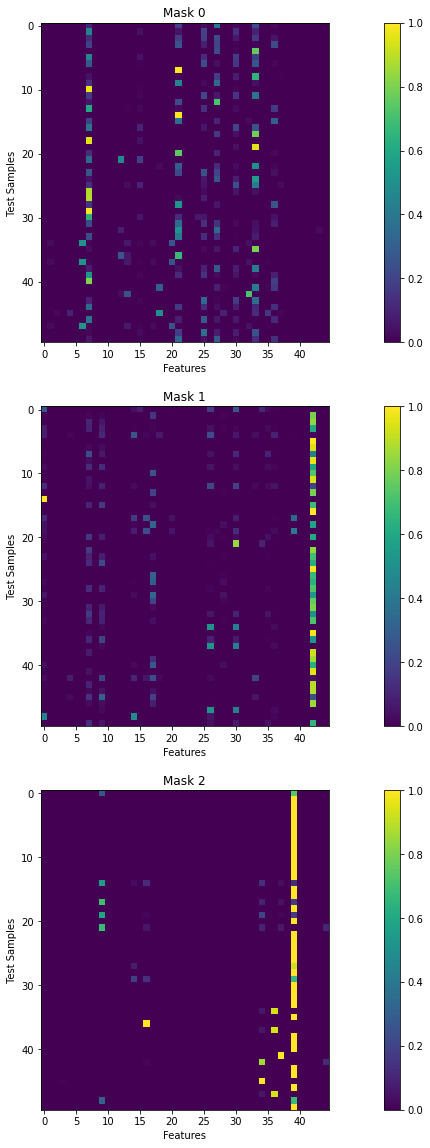}
    \caption{Feature importance using TabNet}
    \label{fig:tabnet_example}
\end{subfigure}
\caption{Visualizing feature importance learnt from entire training data}
\end{figure*}

We formally state our observations as follows:

\begin{itemize}
    \item Learning feature importance from the entire training data, referred to as global methods, can help GNN explainability.
    \item A sliding window where we produce explanations at different levels of model accuracy could be useful.
    \item Graph visualization has its uses, but a \textit{logical explanation} might be more intuitive.
\end{itemize}

A typical explanation subgraph from GNN-Explainer \cite{ying2019gnnexplainer} looks as shown in Figure \ref{fig:gnn_explainer_example}. Similar visualizations can be obtained from other GNN explainability techniques like \cite{ying2019gnnexplainer}, \cite{yuan2020xgnn}, \cite{vu2020pgmexplainer}, \cite{yuan2021explainability}, \cite{schlichtkrull2020interpreting}.

We have in prior works attempted to substantiate GNN model predictions using information retrieval \cite{ganesan2020link}, path ranking \cite{ganesan2020explainable} and reasoner based explanations \cite{bk2021automated}. In \cite{vannur2020data}, we had used a random forest model for post-hoc explanations. In the rest of this extended abstract, we'll present ideas from tabular data.

\section{Feature importance from global methods}



Features selection by looking at the entire training data could be one of the ways to improve both GNN models and explanations which are currently more oriented towards message passing in the local neighborhood and learning the graph structure.

Finding feature importance by looking at all the training data is typical of models trained on tabular data. A typical example from SHAP \cite{lundberg2017unified} on a tabular data prediction is as shown in Figure \ref{fig:shap_example}.

TabNet \cite{TabNet} expands this idea to multiple layers and aggregates the attention information to identify important features. GNN models like GATNE \cite{zhu2019aligraph} have shown how to use attention to improve node representation.

To learn global feature importance with TabNet, we conducted a small experiment by simply converting the graph dataset into tabular data. Our task is to predict if two nodes match and hence in our tabular data we had 146 features (73 for the left entity and 73 for the right) and 67391 entries. Since multiple features consisted entirely of NULL values, they were dropped. Furthermore, we dropped features with less than 500 non-null values. Wholly distinct attributes were dropped since they were unlikely to contribute to the prediction. Even then our dataset ended up with 120 features (60 for each entity).

\begin{figure}[!htb]
    \centering
    \includegraphics[width=\columnwidth]{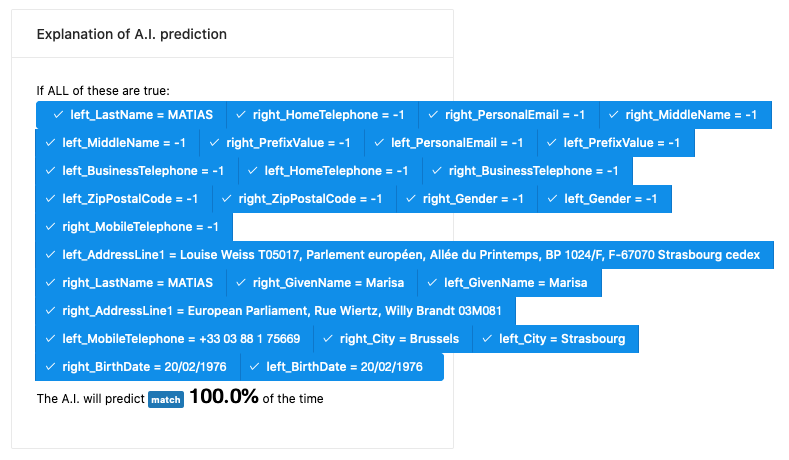}
    \caption{Anchors Explanation at 100\% accuracy}
    \label{fig:anchors_explanation}
\end{figure}
\begin{figure}[!htb]
    \centering
    \includegraphics[width=\columnwidth]{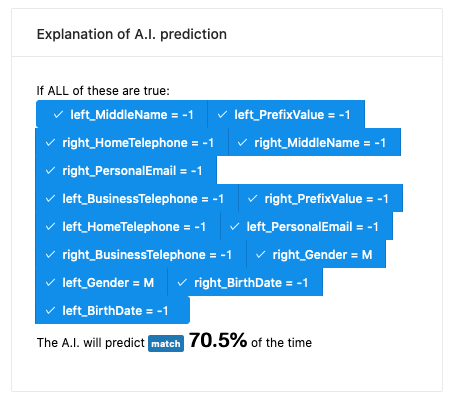}
    \caption{Anchors Explanation at 70.5\% accuracy}
    \label{fig:graph_anchors_explanation}
\end{figure}

\begin{figure*}[!htp]
    \centering
    \includegraphics[height=6cm, width=\linewidth]{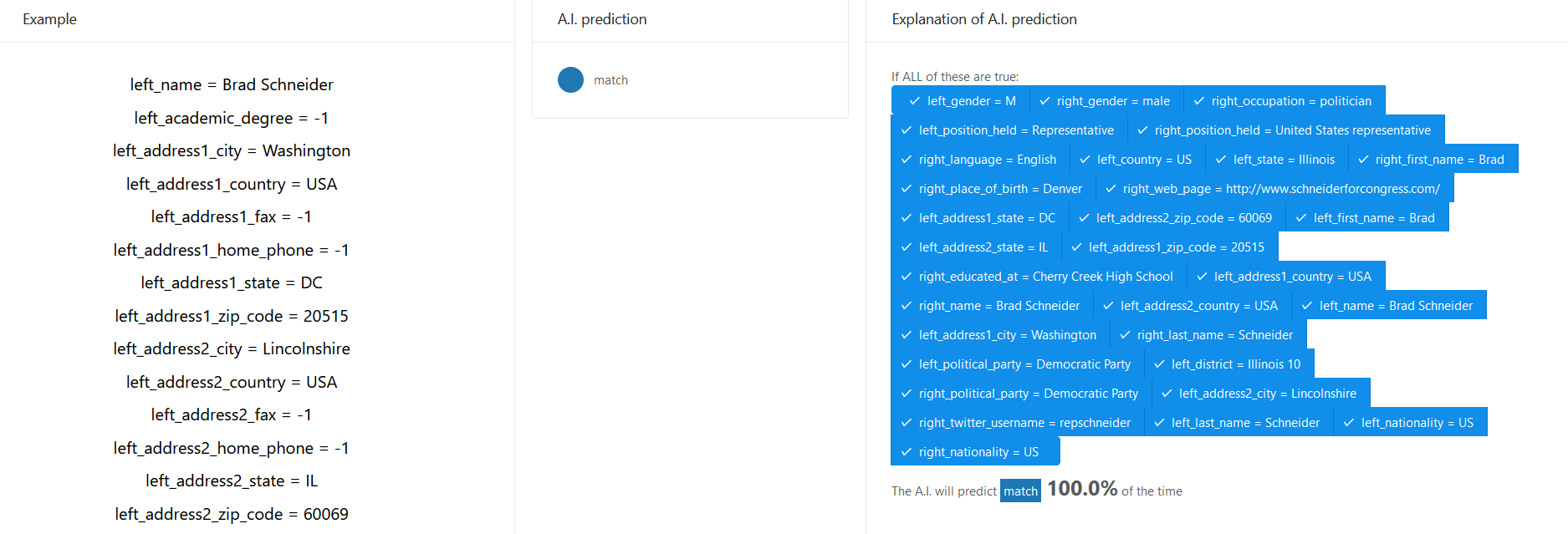}
    \caption{Anchors Explanation with input}
    \label{fig:anchors_full_image}
\end{figure*}

We then trained the TabNet model using the ground truth (match, no\_match) from our graph dataset. The model achieved a test accuracy of 0.8708, a valid\_auc score of 0.93617 and the loss was 0.27686. We were also able to obtain the feature importance information from TabNet using masks as shown in Figure \ref{fig:tabnet_example}. In this figure, X-axis denotes different features and Y axis denotes the samples. Mask i implies feature selection at i-th step. Brighter the color, the higher the feature importance.

Although we did not have human annotations to validate the feature importance as produced by TabNet, we observe that this kind of output could be useful for training better GNN models. Pruning the node features could also lead to better explanations.

\section{Sliding window for explanations}

The next aspect of explainability we find useful is what can be called as the sliding window on accuracy as done in Anchors \cite{ribeiro2018anchors}. Consider the Figures \ref{fig:anchors_explanation} and \ref{fig:graph_anchors_explanation}. Both are explanations from Anchors for the entity matching task. The explanation in Figure \ref{fig:anchors_explanation} is when we ask the explainability technique to assume 100\% model accuracy. Naturally for a model to predict at higher accuracy, it may have to learn lot more features, and hence the explanations may also be quite noisy. If on the other hand, we ask for explanations at a lower accuracy, as shown in Figure \ref{fig:graph_anchors_explanation}, we might be able to understand the predictions better.

Although the explanations on our toy dataset are not that meaningful, we observe that this ability to view explanations at different accuracy levels could be useful in GNN explanations. However, we hasten to add that explanations using methods like GNN Explainer take non-trivial time in some cases, and an interactive setup to consume explanations at different accuracy levels will be lot harder to implement in GNN explanations.

\begin{figure}[htb]
    \centering
    \includegraphics[width=0.98\columnwidth]{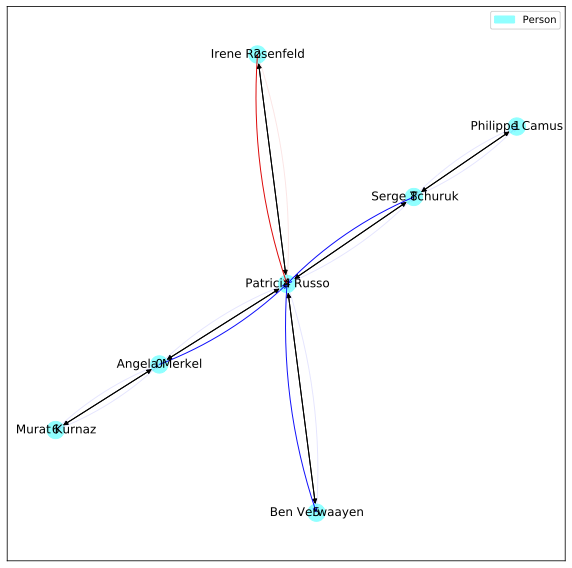}
    \caption{An example GNN Explainer output}
    \label{fig:gnn_explainer_example}
\end{figure}

\begin{figure}[!htp]
    \centering
    \includegraphics[width=\columnwidth]{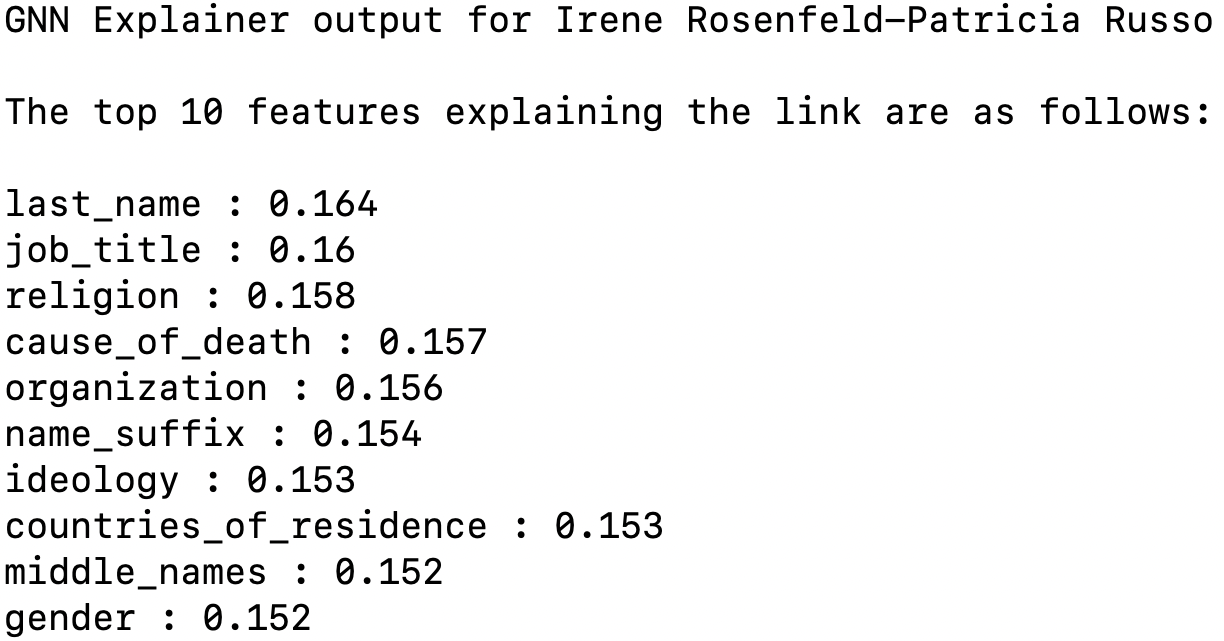}
    \caption{GNN Explainer output as a list of important features}
    \label{fig:feature_mask}
\end{figure}

\begin{figure}[!htp]
    \centering
    \includegraphics[width=0.85\columnwidth]{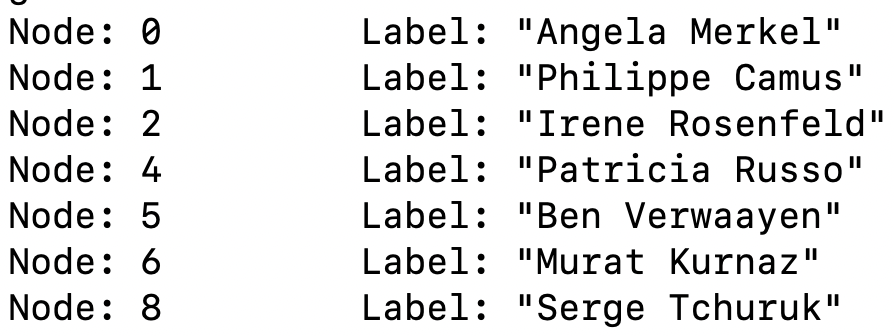}
    \caption{GNN Explainer output with neighbouring nodes}
    \label{fig:node_mask}
\end{figure}

\section{Logical explanations}

Figure \ref{fig:gnn_explainer_example} is the explanation subgraph produced by GNN Explainer \cite{ying2019gnnexplainer}. Many other works have followed this output format but with improvements in the techniques.

Now consider the Anchors explanation in Figure \ref{fig:anchors_full_image}. It displays the input, the prediction and an explanation for a given accuracy rate. We believe these kind of \textit{logical explanations} are more intuitive than graphs in certain cases. Perhaps both the graph visualization and logical explanation could be used together.

Figure \ref{fig:gnn_explainer_example} shows a GNN Explainer output for a link in the TACRED dataset \cite{zhang2017position}. One way to achieve this would be to simply print out the important nodes and features from GNN Explainer like we have done in Figures \ref{fig:feature_mask} and \ref{fig:node_mask}.

\section*{Conclusion}
In this work, we presented three key aspects of explainability techniques from tabular data that could be adopted to Graph Neural Network explanations. We showed example explanations from GNN Explainer, SHAP, TabNet, and Anchors to illustrate our observations.

\section*{Acknowledgements}
This work was done as part of the Global Remote Mentorship initiative of IBM University Relations. We thank Poornima Iyengar, Prof. Balachandra, and Manipal Institute of Technology, Manipal, India for their support. We thank Lokesh Nagalapatti for the discussions.


\bibliography{icml2021}
\bibliographystyle{icml2021}

\end{document}